\documentclass[conference]{IEEEtran}
\usepackage{blindtext, graphicx}
\usepackage{setspace}
\usepackage{algpseudocode}
\usepackage{graphicx}
\usepackage{algorithm}
\usepackage{multirow}
\usepackage{array}
\usepackage{subfigure}
\usepackage{float}
\usepackage{array}
\usepackage{mathtools}
\usepackage{pbox}
\usepackage[english]{babel}
\usepackage[utf8]{inputenc}
\usepackage{fancyhdr}
\usepackage{caption}
\usepackage{amsfonts, amssymb, latexsym}

\newcommand{\indep}{\raisebox{-0.3ex}{\rotatebox{90}{\ensuremath{\models}}}}

\newcommand{\be}{\begin{equation}}
\newcommand{\ee}{\end{equation}}
\newcommand{\bd}{\begin{description}}
\newcommand{\ed}{\end{description}}
\newcommand{\ben}{\begin{enumerate}}
\newcommand{\een}{\end{enumerate}}
\newcommand{\beq}{\begin{quote}}
\newcommand{\eeq}{\end{quote}}
\newcommand{\bi}{\begin{itemize}}
\newcommand{\ei}{\end{itemize}}
\newcommand{\bea}{\begin{eqnarray}}
\newcommand{\eea}{\end{eqnarray}}
\newcommand{\bua}{\begin{eqnarray*}}
\newcommand{\eua}{\end{eqnarray*}}

\newcommand{\ba}{\begin{array}}
\newcommand{\ea}{\end{array}}
\newcommand{\bfig}{\begin{figure}}
\newcommand{\efig}{\end{figure}}
\newcommand{\bc}{\begin{center}}
\newcommand{\ec}{\end{center}}
\newcommand{\bt}{\begin{table}}
\newcommand{\et}{\end{table}}
\newcommand{\btab}{\begin{tabular}}
\newcommand{\etab}{\end{tabular}}

\newcounter{modulenum}
\newcommand{\newspecial}[1]{
        \ifnum\includefigs>0
                \special{#1}
        \fi
}

\begin{document}
%
\title{Latent Variable Discovery Using Dependency Patterns}

\author{\IEEEauthorblockN{...}
\IEEEauthorblockA{Clayton School of Information Technology\\ Monash
  University\\ Clayton, VIC 3800\\ AUSTRALIA\\
Email: Xuhui.Zhang@monash.edu}}

\author{\IEEEauthorblockN{Xuhui Zhang\IEEEauthorrefmark{1},
Kevin B. Korb\IEEEauthorrefmark{2},
Ann E. Nicholson\IEEEauthorrefmark{3}, 
Steven Mascaro\IEEEauthorrefmark{3} 
\IEEEauthorblockA{\IEEEauthorrefmark{1}Clayton School of Information Technology\\ Monash
  University\\ Clayton, VIC 3800\\ AUSTRALIA\\
Bayesian Intelligence\\{\tt www.bayesian-intelligence.com}\\
  Email: Xuhui.Zhang@monash.edu, kbkorb@gmail.com, Ann.Nicholson@monash.edu, sm@voracity.com \\ }
}}

\maketitle

\begin{abstract}
The causal discovery of Bayesian networks is an active and
  important research area, and it is based upon searching the space of causal models for
  those which can best explain a pattern of probabilistic dependencies
  shown in the data. However, some of those dependencies
  are generated by causal structures involving variables which have
  not been measured, i.e., latent variables. Some such patterns of
  dependency ``reveal'' themselves, in that no model based solely upon
  the observed variables can explain them as well as a model using a
  latent variable. That is what latent variable discovery is based
  upon. Here we did a search for finding them systematically, so that they may be
  applied in latent variable discovery in a more rigorous fashion.
\end{abstract}

\begin{IEEEkeywords}
Bayesian networks, Latent variables, causal discovery, probabilistic
dependencies 
\end{IEEEkeywords}

%
\IEEEpeerreviewmaketitle

\section{Introduction}
What enables latent variable discovery is the particular probabilistic dependencies between variables, will typically be representable only by a proper subset of the possible causal models over those variables, and therefore provide evidence in favour of those models and against all the remaining models, as can be
seen in the Bayes factor. Some dependency structures between observed variables will provide evidence favoring latent variable models over fully observed models,\footnote{I.e., models all of whose variables are measured or observed.} because they can explain the dependencies better than any fully observed model. We call such structures
``triggers'' and did a systematically search for them. The result is a clutch of triggers, many of which have not been reported before to our knowledge. These triggers can be
used as a data preprocessing analysis by the main discovery algorithm.

\subsection{Latent Variable Discovery}

Latent variable modeling has a long tradition in causal discovery, beginning
with Spearman's work~\cite{Spear1904} on intelligence
testing. Factor analysis and related methods can be used to posit
latent variables and measure their hypothetical effects. They do not
provide clear means of deciding whether or not latent variables are present
in the first place, however, and in consequence there has been some
controversy about that status of exploratory versus confirmatory
factor analysis. In this regard, causal
discovery methods in AI have the advantage.

\begin{figure}[H]
\begin{center}
\subfigure{
   \label{(a)}
   \includegraphics[width=0.16\textwidth]{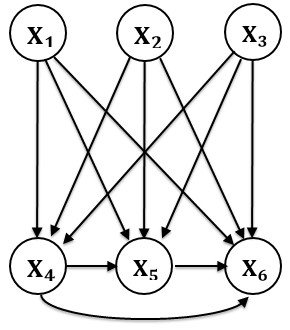}
   } \qquad \qquad      
\subfigure{
    \label{(b)}
   \includegraphics[width=0.16\textwidth]{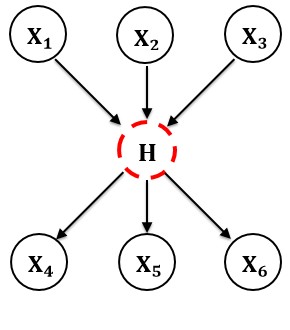}
   }
\end{center}
\caption{An illustration of how introducing a latent variable $H$ can
  simplify a model~\cite{friedman1997learning}.}
\label{fig:friedman}
\end{figure}

One way in which discovery algorithms may find evidence confirmatory
of a latent variable model is in the greater simplicity of such a
model relative to any fully observed model that can represent the data
adequately, as Friedman pointed out using the example in Figure~\ref{fig:friedman}~\cite{friedman1997learning}.

\begin{figure}[H]
\centering
\includegraphics[width=0.25\textwidth]{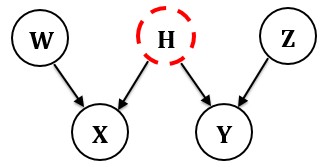}
\caption{One causal structure with four observed variables and one
  latent variable $H$.}
\label{fig:hidden}
\end{figure}

Another advantage for latent variable models is that they can better encode the actual dependencies and independencies in the data. For
example, Figure~\ref{fig:hidden} demonstrates a latent variable model of four
observed variables and one latent variable. If the data support the
independencies $W ~\indep~\{Y,Z\}$ and $Z ~\indep~\{W,X\}$, it is
impossible to construct a network in the observed variables alone that
reflects both of these independencies while also reflecting the
dependencies implied by the d-connections in the latent variable model. It is
this kind of structure which can allow us to infer the existence of
latent variables, i.e., one which constitutes a trigger for latent
variable discovery.

\section{Searching Triggers for Latent Variables}

In this paper, latent variables are typically considered only in scenarios where they
are common causes which having two or more children. As Friedman~\cite{friedman1997learning} points out, a latent variable as a leaf or as a root with only one child would marginalize out without
affecting the distribution over the remaining variables. So too would
a latent variable that mediates only one parent and one child. For simplicity, we also only search for triggers for isolated latent variables rather than multiple latent variables.

We start by enumerating all possible fully observed DAGs in a given number of variables (this step is already super exponential~\cite{robinson1977counting}!. Then it generates all possible d-separating evidence sets. For example, for the four variables $W, X, Y$ and $Z$,
there are eleven evidence sets:\footnote{Note that sets of three or more evidence variables leave nothing left over to separate.}  \[ \phi, \{W\}, \{X\}, \{Y\}, \{Z\},
\{WX\},\] \[\{WY\}, \{WZ\}, \{XY\}, \\ \{XZ\}, \{YZ\}\]

For each fully observed DAG it produces the corresponding dependencies for each
evidence set using the d-separation criterion (i.e., for the four
variables $W, X, Y$ and $Z$, the search produces eleven dependency
matrices). Next, it generates all possible single hidden-variable
models whose latent variable is a common cause of two observed variables. It
then generates all the dependencies between observed variables
for each latent variable model, conditioned upon each evidence
set. The set of dependencies of a latent variable model is a {\bf
  trigger} if and only if these dependency sets cannot be matched by
any fully observed DAG in terms of d-separation.\footnote{In this search, labels (variable
  names) are ignored, of course, since all that matters are the
  dependency structures.}

We ran our search for 3, 4 and 5 observed variables. Any structures
with isolated nodes are not be included. As Table~\ref{tab:number} shows, for three observed variables, we find no trigger, meaning the set of dependencies implied by all possible hidden models can also be found in one or more fully observed models. There are two triggers for four observed variables, the corresponding DAGs are shown in Table~\ref{tab:4var},
together with the corresponding latent variable models. For five
observed variables, we find 57 triggers (see Appendix~\ref{app:triggers}). 

\begin{table} [H]
\begin{center}
\begin{tabular}{|c|c|c|c|}
\hline
\pbox{10cm}{Observed variables} & \quad \small{DAGs} \quad & \quad \pbox{20cm}{Connected DAGs} \quad & \quad \small{Triggers} \quad \\ 
\hline
3      & 6       & 4     &0    \\
4      & 31      & 24    &2    \\
5      & 302     & 268   &57  \\
\hline
\end{tabular}
\end{center}
\caption{Number of triggers found}
\label{tab:number}
\end{table}

\begin{table} [H]
\begin{center}
\begin{tabular}{cc}
\includegraphics[width=0.18\textwidth, height=15mm]{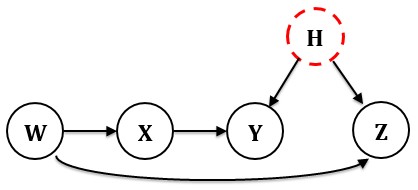} & 
\includegraphics[width=0.18\textwidth, height=15mm]{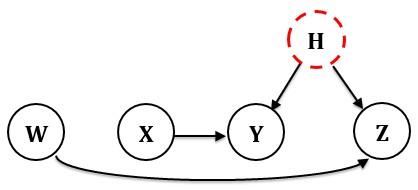} \\ 
\end{tabular}
\end{center}
\caption{The DAGs of the two triggers found for four observed variables.}
\label{tab:4var}
\end{table}

All the dependency structures in the observed variables revealed as
triggers will be better explained with latent variables than
without. While it is not necessary to take triggers into account
explicitly in latent variable discovery, since random structural
mutations combined with standard metrics may well find them, they can
be used to advantage in the discovery process, by focusing it, making
it more efficient and more likely to find the right structure. 

\section{Learning Triggers with Causal Discovery Algorithms}

The most popular causal discovery programs in
general, come from the Carnegie Mellon group and are incorporated into
TETRAD, namely FCI and PC~\cite{tetrad}. Hence, they are the natural
foil against which to compare anything we might produce. Our ultimate
goal is to incorporate latent variable discovery
into a metric-based program. As that is a longer project, here we
report experimental results using an ad hoc arrangement of running the
trigger program as a filter to ordinary PC (Trigger-PC) and comparing
that with the unaltered FCI and PC. Our experimental procedure, briefly, was:

\begin{enumerate}
\item Generate random networks of a given number of variables, with three categories of dependency: weak, medium and strong.
\item Generate artificial data sets using these networks.
\item Optimize the alpha level of the FCI and PC programs using the above.
\item Experimentally test and compare FCI and PC.
\end{enumerate}

The FCI and PC algorithms do not generally return a single DAG, but a
hybrid graph~\cite{spirtes2000causation}. An arc between two nodes may be undirected '\textemdash' or bi-directional '$\leftrightarrow$', which indicates the presence of a latent common cause. Additionally, the graph produced by FCI may contain
'o\textemdash o' or 'o$\rightarrow$'.  The circle represents an
unknown relationship, which means it is not known whether or not an
arrowhead occurs at that end of the arc~\cite{spirtes2000causation}. So, in order to measure how close the models learned by FCI and PC are to the true model, we developed a
special version of the edit distance between graphs (see the Appendix~\ref{editdistanc}).

\subsection{Step one: generate networks with different dependency strengths.}
Genetic algorithms (GAs)~\cite{Russel2003} are commonly applied as a
search algorithm based on an artificial selection process that simulates
biological evolution. Here we used a GA algorithm to find good representative, but
random, graphs with the three levels of desired dependency strengths
between variables: strong, medium and weak. The idea is to test the learning algorithms across different degrees of difficulty in recovering arcs (easy, medium and difficult, respectively).  Mutual information~\cite{pearl1988probabilistic} is used to assess the
strengths of individual arcs in networks.

To make the learning process more efficient, we set the arities for
all nodes in a network to be the same, either two or three. We
randomly initialized all variables' CPT parameters for each individual
graph and used a whole population of 100 individuals. The GA was run
100 generations. We ran the GA for each configuration (number of nodes
and arities) three times, the first two to obtain networks with the
strongest and weakest dependencies between parents and their children
and the third time to obtain networks closest to the average of those
two degrees of strength.

\subsection{Step two: generate artificial datasets.}
All networks with different arc strength levels were used to generate
artificial datasets with sample sizes of 100, 1000 and 10000. We used Netica API~\cite{neticaAPI} to generate random cases. The default
sampling method is called ``Forward Sampling"~\cite{neticasampling}
which is what we used.

\begin{table}[H]
\begin{center}
\begin{tabular}{|c|c|c|c|}
\hline
\pbox{10cm}{Number of \\ observed variables} & \quad \pbox{2cm}{Structure type} \quad & \quad \pbox{3cm}{Number of \\ structures} \quad & \pbox{5cm}{Total number \\ of simulated \\ datasets} \quad  \\ 
\hline
4 &Trigger  & 2       & 36        \\
4 &DAG      & 24      & 432       \\
5 &Trigger  & 57      & 1026      \\
5 &DAG      & 268     & 4824      \\
\hline
\end{tabular}
\end{center}
\caption{Number of simulated datasets}
\label{tab:simulateddatano}
\end{table} 

As Table~\ref{tab:simulateddatano} shows, there are a relatively large
number of simulated datasets. This is due to the different state
numbers, arc strength levels and data sizes. For example, there are 57
trigger structures for 5 observed variables, so there are $57 \times
2 \times 3 \times 3 = 1026$ simulated datasets.

\subsection{Step three: optimize alpha to obtain the shortest edit distance from true models} 

FCI and PC both rely on statistical significance tests to decide whether an arc exists between two variables and on its orientation. They have a default alpha level (of 0.05), but the authors have in the past criticized experimental work
using the default and recommended instead optimizing the alpha level
for the task at hand, so here we do that. The idea is to give the
performance of FCI and PC the benefit of any possible doubt. Given the
artificial data sets generated, we can use FCI and PC with different
values of alpha to learn networks and compare the results to the
models used to generate those data sets. We then used our version of
edit distance between the learned and generating models (see Appendix~\ref{editdistanc}) to find the optimal alpha levels for both algorithms.

We first tried simulated annealing to search for an optimal alpha, but
in the end simply generated sufficiently many random values from the
uniform distribution over the range of [0.0, 0.5]. We evaluated alpha
values for the datasets with 2 states and 3 states separately.

As shown in the following graphs, the average edit distances
between the learned and true models approximate a parabola with a
minimum around 0.1 to 0.2. The results below are specific to
the exact datasets and networks we developed for this experimental
work.

\noindent
1) FCI algorithm 

Results for FCI were broadly similar. In summary, the optimal alphas found for the above cases (in the same
order) were: 0.12206, 0.19397, 0.20862 and 0.12627.

\begin{itemize}
   \item Number of observed variables: 4 \\
        Datasets: 2 state DAG structure simulated dataset \\
        Number of datasets: 24*9 = 216
       \begin{figure} [h]
       \centering
       \includegraphics[width=0.4\textwidth]{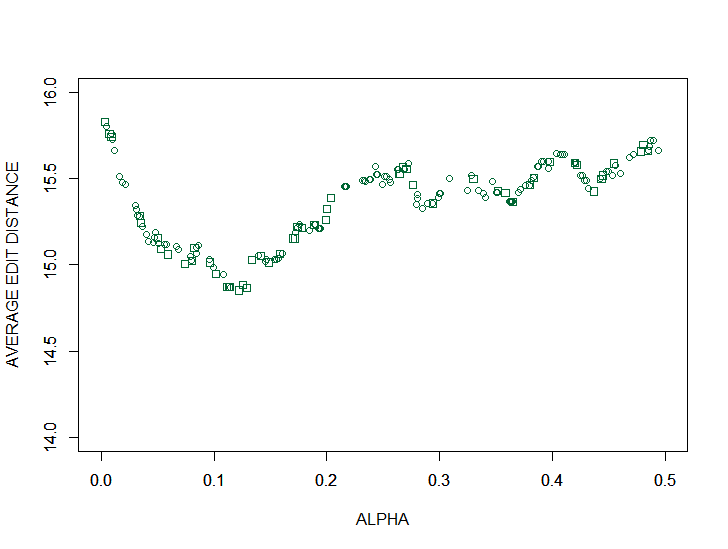}
       \label{fig:fci4var2state}
       \end{figure}
       
       Results: \\
       Minimum average edit distance: 14.85185 \\
       Maximum average edit distance: 15.82870 \\
       Mean average edit distance: 15.37168    \\
       Best Alpha: 0.12206 \\
       95\% confidence level: 0.03242 \\
       95\% confidence interval: (15.37168-0.03242, 15.37168+0.03242) \\

   \item Number of observed variables: 4 \\
        Datasets: 3 state DAG structure simulated dataset \\
        Number of datasets: 24*9 = 216

       \begin{figure} [H]
       \centering
       \includegraphics[width=0.4\textwidth]{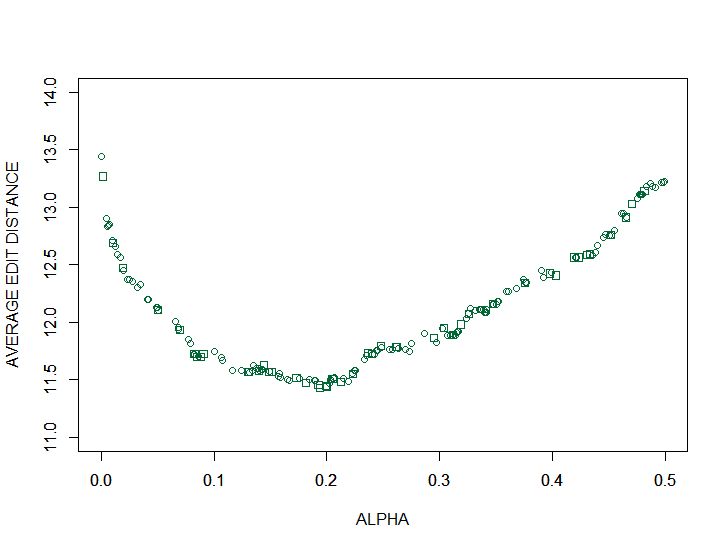}
       \label{fig:fci4var3state}
       \end{figure}
       
       Results: \\
       Minimum average edit distance: 11.42593 \\
       Maximum average edit distance: 13.43981 \\
       Mean average edit distance: 12.09355    \\
       Best Alpha: 0.19397 \\
       95\% confidence level: 0.07259 \\
       95\% confidence interval: (12.09355-0.07259, 12.09355+0.07259) \\

      \item Number of observed variables: 5 \\
        Datasets: 2 state DAG structure simulated dataset \\
        Number of datasets: 268*9 = 2412

       \begin{figure} [H]
       \centering
       \includegraphics[width=0.4\textwidth]{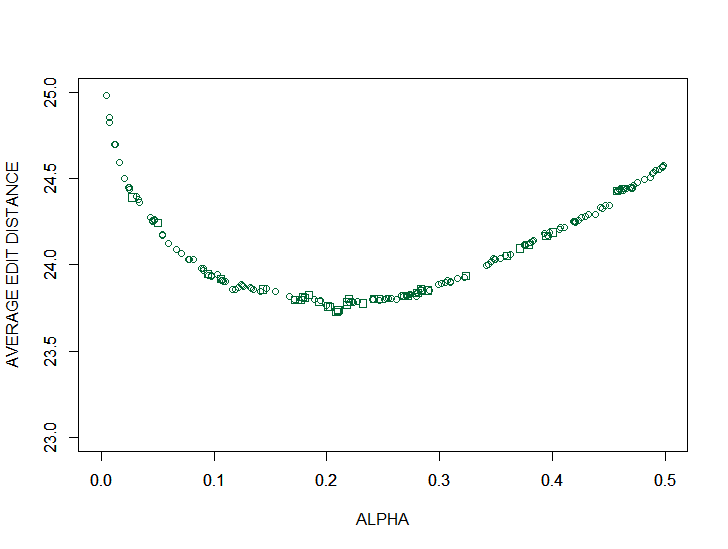}
       \label{fig:fci5var2state}
       \end{figure}
       
       Results: \\
       Minimum average edit distance: 23.72844 \\
       Maximum average edit distance: 24.98466 \\
       Mean average edit distance: 24.08980    \\
       Best Alpha: 0.20862 \\
       95\% confidence level: 0.03880 \\
       95\% confidence interval: (24.08980-0.03880, 24.08980-0.03880) \\
       
       \item Number of observed variables: 5 \\
        Datasets: 3 state DAG structure simulated dataset \\
        Number of datasets: 268*9 = 2412

       \begin{figure} [H]
       \centering
       \includegraphics[width=0.4\textwidth]{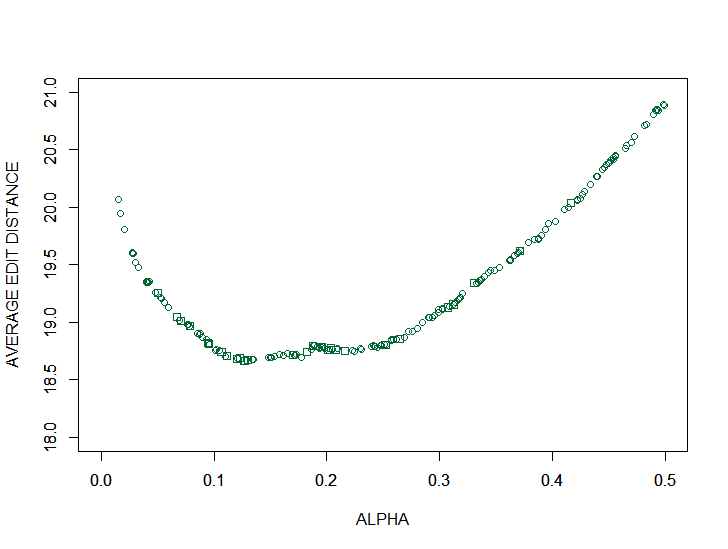}
       \label{fig:fci5var3state}
       \end{figure}
       
       Results: \\
       Minimum average edit distance: 18.66501 \\
       Maximum average edit distance: 20.89221 \\
       Mean average edit distance: 19.32648    \\
       Best Alpha: 0.12627 \\
       95\% confidence level: 0.08906 \\
       95\% confidence interval: (19.32648-0.08906, 19.32648-0.08906) \\

\end{itemize}

\noindent
2) PC algorithm 

Results for PC were quite similar. In
summary, the optimal alphas found for the above cases (in the same
order) were: 0.12268, 0.20160, 0.20676 and 0.13636.

\subsection{Step four: compare the learned models with true model.}

Finally, we were ready to test FCI and PC on the artificial datasets
of trigger (i.e., latent variable) and DAG structures. Artificial
datasets generated by trigger structures were used to determine True
Positive (TP) and False Negative (FN) results (i.e., finding the real
latent and missing the real latent, respectively), while the datasets
of (fully observed) DAG structures were used for False Positive (FP)
and True Negative (TN) results. Assume the latent variable in every
trigger structure is the parent of node A and B, we used the following
definitions:

\begin{itemize}
\item TP: The learned model has a bi-directional arc between A and B. 
\item FN: The learned model lacks a bi-directional arc between A and B.
\item TN: The learned model has no bi-directional arcs.
\item FP: The learned model has one or more bi-directional arcs.
\end{itemize} 

We tested the FCI and PC algorithms on different datasets with their
corresponding optimized alphas. We do not report confidence intervals
or significance tests between different algorithms under different
conditions, since the cumulative results over 6,318 datasets suffices
to tell the comparative story.

The following tables show the confusion matrix summing over all
datasets (see Appendix~\ref{app:conmatrix} for more detailed results):

\begin{table}[H]
\begin{center}
\begin{tabular}{|c|c|c|c|c|}
\hline
& \multicolumn{2}{|c|}{FCI} & \multicolumn{2}{|c|}{PC}\\
\hline
& \quad Latent &\quad No Latent &\quad  Latent &\quad No Latent\\ 
\hline
 Positive &235 &981 &226 &819     \\
\hline
 Negative &827 &4275 &836 &4437     \\
\hline
\end{tabular}
\end{center}
\label{tab:confusionmatrix}
\end{table} 

With corresponding optimal alpha, the FCI's predictive accuracy was
0.71 (rounding off), its precision 0.19, its recall 0.22 and its false
positive rate 0.19.  The predictive accuracy for PC was 0.74, the
precision was 0.22, its recall 0.21 and the false positive rate 
was 0.16.

We also did the same tests using the default alpha of 0.05. The
results are shown as follow (see Appendix~\ref{app:conmatrix} for more detailed
results):

\begin{table}[H]
\begin{center}
\begin{tabular}{|c|c|c|c|c|}
\hline
& \multicolumn{2}{|c|}{FCI} & \multicolumn{2}{|c|}{PC}\\
\hline
& \quad Latent &\quad No Latent &\quad  Latent &\quad No Latent\\ 
\hline
 Positive &211 &767 &205 &615     \\
\hline
 Negative &851 &4489 &857 &4641     \\
\hline
\end{tabular}
\end{center}
\label{tab:confusionmatrix2}
\end{table} 

With alpha of 0.05, the FCI's predictive accuracy was 0.74, its
precision 0.22, its recall 0.19 and its false positive rate 0.15.
PC's predictive accuracy was 0.77, the precision was 0.25, its
recall 0.19 and the false positive rate was 0.12.

As we can see from the results, the performance of FCI and PC are
quite similar. Neither are finding the majority of latent variables
actually there, but both are at least showing moderate false positive
rates. Arguably, false positives are a worse offence than false
negatives, since false negatives leave the causal discovery process no
worse off than an algorithm that ignores latents, whereas a false
positive will positively mislead the causal discovery process.

\section{Applying Triggers in Causal Discovery}

\subsection{An extension of PC algorithm (Trigger-PC)}

We implemented triggers as a data filter into PC, yielding Trigger-PC, and see how well it would work. If Trigger-PC finds a trigger pattern in the data, then it returns that trigger
structure, otherwise it returns whatever structure the PC algorithm
returns, while replacing any incorrect bi-directed arcs by undirected
arcs. So the Trigger-PC algorithm (see Algorithm~\ref{alg:Trigger-PC})
gives us a more specific latent model structure and, as we shall see,
has fewer false positives.

\begin{algorithm}[ht]
  \caption{Trigger-PC Algorithm}
\label{alg:Trigger-PC}
  \begin{algorithmic}[1]
    \State Let $D$ be the test dataset;
    \State Let $D\_labels$ be the variable labels in $D$;
    \State Let $Triggers$ be the triggers given the number of variables in $D$;
    \State Perform conditional significant tests to get the dependency pattern $P$ in $D$;
    \State Let $matchTrigger = false$;
    \State Let $T$ be an empty DAG;
    \State Let $label\_assignments$ be all possible label assignments of $D\_label$;
    \For{each $trigger$ in $Triggers$}
      \State Let $t$ be the unlabeled DAG represented by $trigger$;
      \For{each $label\_assignment$ in $label\_assignments$} 
        \State Assign $label\_assignment$ to $t$, yield $t\_labeled$;
        \State Generate dependency pattern of $t\_labeled$, yield $t\_pattern$;
        \If{$P$ matches $t\_pattern$}
          \State $matchTrigger = true$;
          \State $T = t\_labeled$;
          \State break;
        \EndIf  
      \EndFor
      \If{$matchTrigger = true$}
      \State break;
      \EndIf
    \EndFor
    \If{$matchTrigger = true$}
      \State Output $T$;
    \EndIf
    \If{$matchTrigger = false$}
      \State Run PC Algorithm with input dataset $D$;
      \State Let $PC\_result$ be the result structure produced by PC Algorithm;
      \If{there is any bi-directed arcs in $PC\_result$}
         \State Replace all bi-directed arcs by undirected arcs, yield $PC\_result*$;
         \State Output $PC\_result*$;
      \EndIf
    \EndIf
  \end{algorithmic}
\end{algorithm}

We tested our Trigger-PC algorithm with the alpha optimized for the PC
algorithm. The resultant confusion matrix was (see Appendix~\ref{app:conmatrix} for more details):

\begin{table}[H]
\begin{center}
\begin{tabular}{|c|c|c|}
\hline
& \multicolumn{2}{|c|}{trigger-PC}\\
\hline
& \quad Latent &\quad No Latent \\ 
\hline
 Positive &30 &3 \\
\hline
 Negative &1032 &5253 \\
\hline
\end{tabular}
\end{center}
\label{tab:triggerpc-confusionmatrix}
\end{table}

Trigger-PC's predictive accuracy was 0.84, its precision 0.91, its recall 0.03 and the false positive rate 0.0006. We can see that Trigger-PC is finding far fewer latents than either PC or FCI, but when it asserts their existence we can have much greater confidence in the claim. As we indicated above, avoiding false positives, while having at least some true positives, appears to be the more important goal in latent variable discovery.

As before, we also tried the default 0.05 alpha in Trigger-PC, with the results (see mode details in Appendix~\ref{app:conmatrix}):

\begin{table}[H]
\begin{center}
\begin{tabular}{|c|c|c|}
\hline
& \multicolumn{2}{|c|}{trigger-PC}\\
\hline
& \quad Latent &\quad No Latent \\ 
\hline
 Positive &35 &4 \\
\hline
 Negative &1027 &5252 \\
\hline
\end{tabular}
\end{center}
\label{tab:triggerpc-confusionmatrix2}
\end{table}

And, again, these results are only slightly different.
With alpha of 0.05, the trigger-PC's predictive accuracy was 0.84, its precision 0.90, its recall 0.03 and the false positive rate 0.0008.

\section{Conclusion}

We have presented the first systematic search algorithm to discover
and report latent variable triggers: conditional probability
structures that are better explained by latent variable models than by
any DAG constructed from the observed variables alone.  For simplicity
and efficiency, we have limited this to looking for single latent
variables at a time, although that restriction can be removed.  We
have also applied this latent discovery algorithm directly in an
existing causal discovery algorithm and compared the results to
existing algorithms which discover latents using different
methods. The results are certainly different and arguably superior.
Our next step will be to implement this approach within a metric-based
causal discovery program.

\bibliographystyle{ieeetran}

\appendices
\onecolumn

\section{57 triggers for latent variable models found for five observed variables}
\label{app:triggers}

\begin{figure} [H]
\centering
\includegraphics[width=0.9\textwidth]{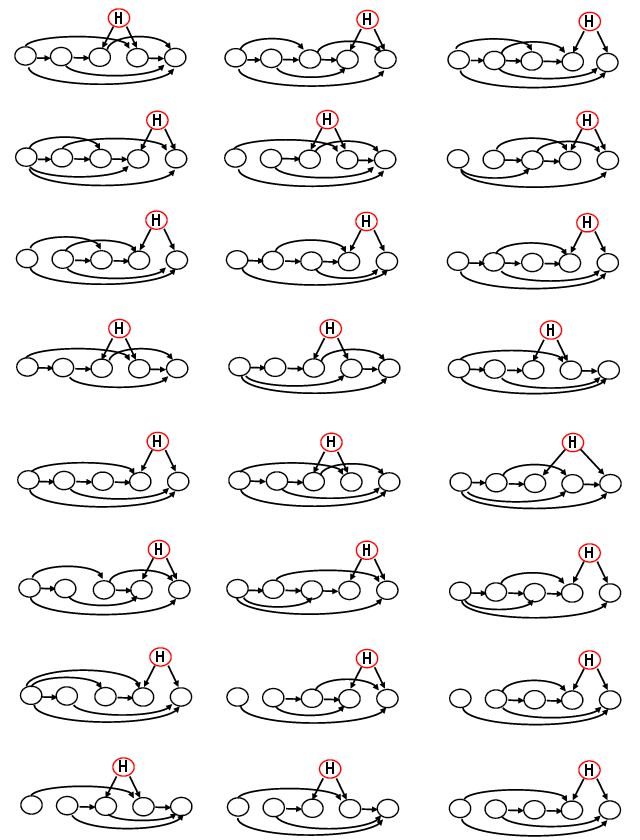} 
\label{fig:triggermodels1}
\end{figure}

\begin{figure} [H]
\centering
\includegraphics[width=0.9\textwidth]{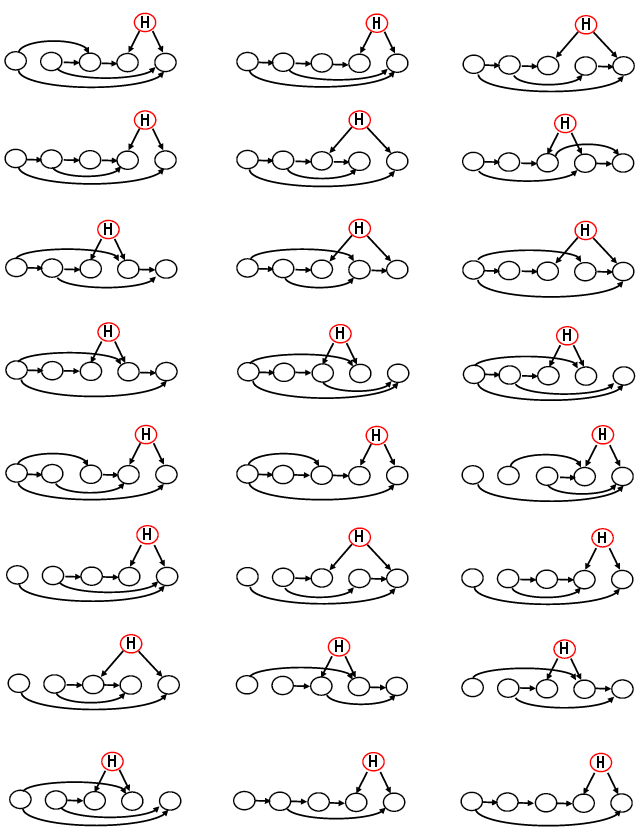} 
\label{fig:triggermodels2}
\end{figure}

\begin{figure} [H]
\centering
\includegraphics[width=0.9\textwidth]{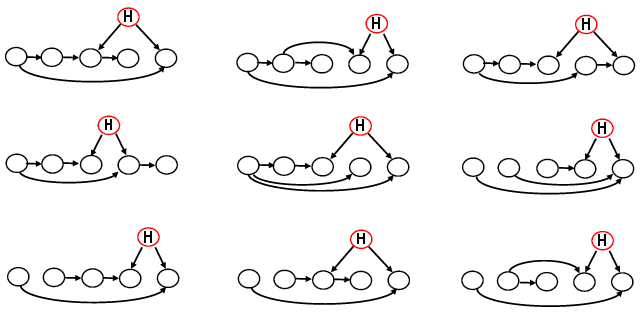} 
\label{fig:triggermodels3}
\end{figure}

\section{Edit distance used in causal model results}
\label{editdistanc}
Edit distance between different type of arcs (between variable A and B) produced by PC algorithm and true arc type:
\begin{table}[H]
\centering
\label{tab:pc_editdistance}
\begin{tabular}{|c|c|c|c|c|c|c|}
\hline
True arc                        &A                      &B                       &Learned arc     &A     &B     &Distance  \\ \hline
\multirow{5}{*}{A $\rightarrow$ B}    & \multirow{5}{*}{Tail} & \multirow{5}{*}{Arrow} &A\textemdash B                   &Tail  &Tail  &2 \\
\cline{4-7} 
                                      &                       &                        &A $\rightarrow$ B     &Tail  &Arrow &0 \\ 
\cline{4-7} 
                                      &                       &                        &A $\leftarrow$ B      &Arrow &Tail  &4 \\ 
\cline{4-7} 
                                      &                       &                        &A $\leftrightarrow$ B &Arrow &Arrow &2 \\ \cline{4-7} 
                                      &                       &                        &null                  &null  &null  &6 \\ \hline
\multicolumn{7}{|l|}{}                                                \\                                                          \hline
\multirow{5}{*}{A $\leftrightarrow$ B}& \multirow{5}{*}{Arrow} & \multirow{5}{*}{Arrow} &A\textemdash B                   &Tail  &Tail  &4 \\
\hline
\end{tabular}
\end{table}

Edit distance between different type of arcs (between variable A and B) produced by FCI algorithm and true arc type:
\begin{table}[H]
\centering
\label{tab:fci_editdistance}
\begin{tabular}{|c|c|c|c|c|c|c|}
\hline
True arc                        &A                      &B                       &Learned arc     &A     &B     &Distance  \\ \hline
\multirow{5}{*}{A $\rightarrow$ B}    & \multirow{5}{*}{Tail} & \multirow{5}{*}{Arrow} &A\textemdash B                   &Tail  &Tail  &2 \\
\cline{4-7} 
                                      &                       &                        &A $\rightarrow$ B      &Tail  &Arrow &0 \\ 
\cline{4-7} 
                                      &                       &                        &A $\leftarrow$ B      &Arrow  &Tail &4 \\ 
\cline{4-7} 
                                      &                       &                        &A o\textemdash B    &Circle &Tail  &3 \\
\cline{4-7} 
                                      &                       &                        &A \textemdash o B    &Tail &Circle  &1 \\
\cline{4-7} 
                                      &                       &                        &A o$\rightarrow$ B    &Circle &Arrow  &1 \\
\cline{4-7} 
                                      &                       &                        &A $\leftarrow$o B     &Arrow &Circle  &3 \\
\hline
\end{tabular}
\end{table}

\section{Confusion matrix}
\label{app:conmatrix}

1) Results of FCI algorithm (with optimized alpha)

\begin{itemize}

\item 4 observed variables with 2 state each simulated data \\

Alpha: 0.12206 \\

Maximum arc strength simulated data:

\begin{table}[H]
\centering
\begin{tabular}{|c|c|c|c|c|c|c|}
\hline
 Data case number& \multicolumn{2}{c|}{100} & \multicolumn{2}{c|}{1000} & \multicolumn{2}{c|}{10000} \\ \hline
                 &Latent         &No latent            &Latent           &No latent           &Latent           &No latent            \\ \hline
        Positive &0          &0             &1            &0            &0            &0             \\ \hline
        Negative &2         &24             &1           &24            &2           &24             \\ \hline
\end{tabular}
\end{table}

Medium arc strength simulated data:

\begin{table}[H]
\centering
\begin{tabular}{|c|c|c|c|c|c|c|}
\hline
 Data case number& \multicolumn{2}{c|}{100} & \multicolumn{2}{c|}{1000} & \multicolumn{2}{c|}{10000} \\ \hline
                 &Latent         &No latent            &Latent           &No latent           &Latent           &No latent            \\ \hline
        Positive &0          &0             &2            &2           &1            &0             \\ \hline
        Negative &2         &24             &0           &22            &1           &24             \\ \hline
\end{tabular}
\end{table}

Minimum arc strength simulated data:

\begin{table}[H]
\centering
\begin{tabular}{|c|c|c|c|c|c|c|}
\hline
 Data case number& \multicolumn{2}{c|}{100} & \multicolumn{2}{c|}{1000} & \multicolumn{2}{c|}{10000} \\ \hline
                 &Latent         &No latent            &Latent           &No latent           &Latent           &No latent            \\ \hline
        Positive &0          &0             &0            &0            &0            &2             \\ \hline
        Negative &2         &24             &2           &24            &2           &22             \\ \hline
\end{tabular}
\end{table}

\item 4 observed variables with 3 state each simulated data \\

Alpha: 0.19397 \\

Maximum arc strength simulated data:

\begin{table}[H]
\centering
\begin{tabular}{|c|c|c|c|c|c|c|}
\hline
 Data case number& \multicolumn{2}{c|}{100} & \multicolumn{2}{c|}{1000} & \multicolumn{2}{c|}{10000} \\ \hline
                 &Latent         &No latent            &Latent           &No latent           &Latent           &No latent            \\ \hline
        Positive &0          &0             &1            &0            &2            &1             \\ \hline
        Negative &2        &24             &1           &24            &0           &23             \\ \hline
\end{tabular}
\end{table}

Medium arc strength simulated data:

\begin{table}[H]
\centering
\begin{tabular}{|c|c|c|c|c|c|c|}
\hline
 Data case number& \multicolumn{2}{c|}{100} & \multicolumn{2}{c|}{1000} & \multicolumn{2}{c|}{10000} \\ \hline
                 &Latent         &No latent            &Latent           &No latent           &Latent           &No latent            \\ \hline
        Positive &0          &4             &1            &1            &2            &0             \\ \hline
        Negative &2         &20             &1           &23            &0          &24             \\ \hline
\end{tabular}
\end{table}

Minimum arc strength simulated data:

\begin{table}[H]
\centering
\begin{tabular}{|c|c|c|c|c|c|c|}
\hline
 Data case number& \multicolumn{2}{c|}{100} & \multicolumn{2}{c|}{1000} & \multicolumn{2}{c|}{10000} \\ \hline
                 &Latent         &No latent            &Latent           &No latent           &Latent           &No latent            \\ \hline
        Positive &0          &1             &0            &2            &0            &4             \\ \hline
        Negative &2         &23             &2           &22            &2          &20             \\ \hline
\end{tabular}
\end{table}

\item 5 observed variables with 2 state each simulated data \\

Alpha: 0.20862 \\

Maximum arc strength simulated data:

\begin{table}[H]
\centering
\begin{tabular}{|c|c|c|c|c|c|c|}
\hline
 Data case number& \multicolumn{2}{c|}{100} & \multicolumn{2}{c|}{1000} & \multicolumn{2}{c|}{10000} \\ \hline
                 &Latent         &No latent            &Latent           &No latent           &Latent           &No latent            \\ \hline
        Positive &1          &38             &3            &43            &12            &30             \\ \hline
        Negative &56         &230             &54           &225            &45           &238             \\ \hline
\end{tabular}
\end{table}

Medium arc strength simulated data:

\begin{table}[H]
\centering
\begin{tabular}{|c|c|c|c|c|c|c|}
\hline
 Data case number& \multicolumn{2}{c|}{100} & \multicolumn{2}{c|}{1000} & \multicolumn{2}{c|}{10000} \\ \hline
                 &Latent         &No latent            &Latent           &No latent           &Latent           &No latent            \\ \hline
        Positive &1          &52             &16            &72            &34            &57             \\ \hline
        Negative &56         &216             &41           &196            &23           &211            \\ \hline
\end{tabular}
\end{table}

Minimum arc strength simulated data:

\begin{table}[H]
\centering
\begin{tabular}{|c|c|c|c|c|c|c|}
\hline
 Data case number& \multicolumn{2}{c|}{100} & \multicolumn{2}{c|}{1000} & \multicolumn{2}{c|}{10000} \\ \hline
                 &Latent         &No latent            &Latent           &No latent           &Latent           &No latent            \\ \hline
        Positive &0          &7             &0            &26            &0            &41             \\ \hline
        Negative &57         &261             &57           &242            &57           &227             \\ \hline
\end{tabular}
\end{table}

\item 5 observed variables with 3 state each simulated data \\

Alpha: 0.12627 \\

Maximum arc strength simulated data:

\begin{table}[H]
\centering
\begin{tabular}{|c|c|c|c|c|c|c|}
\hline
 Data case number& \multicolumn{2}{c|}{100} & \multicolumn{2}{c|}{1000} & \multicolumn{2}{c|}{10000} \\ \hline
                 &Latent         &No latent            &Latent           &No latent           &Latent           &No latent            \\ \hline
        Positive &6          &82             &26            &59            &51            &16             \\ \hline
        Negative &51         &186             &31           &209            &6           &252             \\ \hline
\end{tabular}
\end{table}

Medium arc strength simulated data:

\begin{table}[H]
\centering
\begin{tabular}{|c|c|c|c|c|c|c|}
\hline
 Data case number& \multicolumn{2}{c|}{100} & \multicolumn{2}{c|}{1000} & \multicolumn{2}{c|}{10000} \\ \hline
                 &Latent         &No latent            &Latent           &No latent           &Latent           &No latent            \\ \hline
        Positive &1          &104             &19           &98            &46            &15             \\ \hline
        Negative &56        &164             &38           &170           &11           &253             \\ \hline
\end{tabular}
\end{table}

Minimum arc strength simulated data:

\begin{table}[H]
\centering
\begin{tabular}{|c|c|c|c|c|c|c|}
\hline
 Data case number& \multicolumn{2}{c|}{100} & \multicolumn{2}{c|}{1000} & \multicolumn{2}{c|}{10000} \\ \hline
                 &Latent         &No latent            &Latent           &No latent           &Latent           &No latent            \\ \hline
        Positive &0          &6             &1            &80            &8            &138             \\ \hline
        Negative &57         &262             &56           &168            &49           &130             \\ \hline
\end{tabular}
\end{table}

\end{itemize}

2) Results of PC algorithm (with optimized alpha)

\begin{itemize}

\item 4 observed variables with 2 state each simulated data \\

Alpha: 0.12268 \\

Maximum arc strength simulated data:

\begin{table}[H]
\centering
\begin{tabular}{|c|c|c|c|c|c|c|}
\hline
 Data case number& \multicolumn{2}{c|}{100} & \multicolumn{2}{c|}{1000} & \multicolumn{2}{c|}{10000} \\ \hline
                 &Latent         &No latent            &Latent           &No latent           &Latent           &No latent            \\ \hline
        Positive &0          &0             &1            &0            &0            &0             \\ \hline
        Negative &2         &24             &1           &24            &2           &24             \\ \hline
\end{tabular}
\end{table}

Medium arc strength simulated data:

\begin{table}[H]
\centering
\begin{tabular}{|c|c|c|c|c|c|c|}
\hline
 Data case number& \multicolumn{2}{c|}{100} & \multicolumn{2}{c|}{1000} & \multicolumn{2}{c|}{10000} \\ \hline
                 &Latent         &No latent            &Latent           &No latent           &Latent           &No latent            \\ \hline
        Positive &0          &0             &2            &1            &1            &0             \\ \hline
        Negative &2         &24             &0           &23            &1           &24             \\ \hline
\end{tabular}
\end{table}

Minimum arc strength simulated data:

\begin{table}[H]
\centering
\begin{tabular}{|c|c|c|c|c|c|c|}
\hline
 Data case number& \multicolumn{2}{c|}{100} & \multicolumn{2}{c|}{1000} & \multicolumn{2}{c|}{10000} \\ \hline
                 &Latent         &No latent            &Latent           &No latent           &Latent           &No latent            \\ \hline
        Positive &0          &0             &0            &0            &0            &2             \\ \hline
        Negative &2         &24             &2           &24            &2          &22             \\ \hline
\end{tabular}
\end{table}

\item 4 observed variables with 3 state each simulated data \\

Alpha: 0.20160 \\

Maximum arc strength simulated data:

\begin{table}[H]
\centering
\begin{tabular}{|c|c|c|c|c|c|c|}
\hline
 Data case number& \multicolumn{2}{c|}{100} & \multicolumn{2}{c|}{1000} & \multicolumn{2}{c|}{10000} \\ \hline
                 &Latent         &No latent            &Latent           &No latent           &Latent           &No latent            \\ \hline
        Positive &0          &0             &1            &0            &2            &1             \\ \hline
        Negative &2         &24             &1           &24            &0           &23             \\ \hline
\end{tabular}
\end{table}

Medium arc strength simulated data:

\begin{table}[H]
\centering
\begin{tabular}{|c|c|c|c|c|c|c|}
\hline
 Data case number& \multicolumn{2}{c|}{100} & \multicolumn{2}{c|}{1000} & \multicolumn{2}{c|}{10000} \\ \hline
                 &Latent         &No latent            &Latent           &No latent           &Latent           &No latent            \\ \hline
        Positive &0          &2             &1            &1            &2            &0             \\ \hline
        Negative &2         &22             &1           &23            &0           &24             \\ \hline
\end{tabular}
\end{table}

Minimum arc strength simulated data:

\begin{table}[H]
\centering
\begin{tabular}{|c|c|c|c|c|c|c|}
\hline
 Data case number& \multicolumn{2}{c|}{100} & \multicolumn{2}{c|}{1000} & \multicolumn{2}{c|}{10000} \\ \hline
                 &Latent         &No latent            &Latent           &No latent           &Latent           &No latent            \\ \hline
        Positive &0          &0             &0            &2            &0            &4             \\ \hline
        Negative &2         &24             &2           &22            &2           &20             \\ \hline
\end{tabular}
\end{table}

\item 5 observed variables with 2 state each simulated data \\

Alpha: 0.20676 \\

Maximum arc strength simulated data:

\begin{table}[H]
\centering
\begin{tabular}{|c|c|c|c|c|c|c|}
\hline
 Data case number& \multicolumn{2}{c|}{100} & \multicolumn{2}{c|}{1000} & \multicolumn{2}{c|}{10000} \\ \hline
                 &Latent         &No latent            &Latent           &No latent           &Latent           &No latent            \\ \hline
        Positive &1          &38             &4            &29            &11            &16             \\ \hline
        Negative &56         &230             &53           &239            &46           &252             \\ \hline
\end{tabular}
\end{table}

Medium arc strength simulated data:

\begin{table}[H]
\centering
\begin{tabular}{|c|c|c|c|c|c|c|}
\hline
 Data case number& \multicolumn{2}{c|}{100} & \multicolumn{2}{c|}{1000} & \multicolumn{2}{c|}{10000} \\ \hline
                 &Latent         &No latent            &Latent           &No latent           &Latent           &No latent            \\ \hline
        Positive &1          &49             &15            &52            &30            &40             \\ \hline
        Negative &56         &219             &42           &216           &27           &228             \\ \hline
\end{tabular}
\end{table}

Minimum arc strength simulated data:

\begin{table}[H]
\centering
\begin{tabular}{|c|c|c|c|c|c|c|}
\hline
 Data case number& \multicolumn{2}{c|}{100} & \multicolumn{2}{c|}{1000} & \multicolumn{2}{c|}{10000} \\ \hline
                 &Latent         &No latent            &Latent           &No latent           &Latent           &No latent            \\ \hline
        Positive &0          &7             &0            &23            &0            &41             \\ \hline
        Negative &57         &261             &57           &245            &57           &227             \\ \hline
\end{tabular}
\end{table}

\item 5 observed variables with 3 state each simulated data \\

Alpha: 0.13636 \\

Maximum arc strength simulated data:

\begin{table}[H]
\centering
\begin{tabular}{|c|c|c|c|c|c|c|}
\hline
 Data case number& \multicolumn{2}{c|}{100} & \multicolumn{2}{c|}{1000} & \multicolumn{2}{c|}{10000} \\ \hline
                 &Latent         &No latent            &Latent           &No latent           &Latent           &No latent            \\ \hline
        Positive &6          &73             &24            &43            &49            &8             \\ \hline
        Negative &51         &195             &33           &225            &8           &260             \\ \hline
\end{tabular}
\end{table}

Medium arc strength simulated data:

\begin{table}[H]
\centering
\begin{tabular}{|c|c|c|c|c|c|c|}
\hline
 Data case number& \multicolumn{2}{c|}{100} & \multicolumn{2}{c|}{1000} & \multicolumn{2}{c|}{10000} \\ \hline
                 &Latent         &No latent            &Latent           &No latent           &Latent           &No latent            \\ \hline
        Positive &1          &85            &19           &74            &46            &13             \\ \hline
        Negative &56         &183            &38           &194            &11           &255             \\ \hline
\end{tabular}
\end{table}

Minimum arc strength simulated data:

\begin{table}[H]
\centering
\begin{tabular}{|c|c|c|c|c|c|c|}
\hline
 Data case number& \multicolumn{2}{c|}{100} & \multicolumn{2}{c|}{1000} & \multicolumn{2}{c|}{10000} \\ \hline
                 &Latent         &No latent            &Latent           &No latent           &Latent           &No latent            \\ \hline
        Positive &0          &7             &1            &83            &8            &125             \\ \hline
        Negative &57         &261             &56           &185            &49           &143             \\ \hline
\end{tabular}
\end{table}

\end{itemize}


3) Results of trigger-PC algorithm (with optimized alpha)

\begin{itemize}

\item 4 observed variables with 2 state each simulated data \\

Alpha: 0.12268 \\

Maximum arc strength simulated data:

\begin{table}[H]
\centering
\begin{tabular}{|c|c|c|c|c|c|c|}
\hline
 Data case number& \multicolumn{2}{c|}{100} & \multicolumn{2}{c|}{1000} & \multicolumn{2}{c|}{10000} \\ \hline
                 &Latent         &No latent            &Latent           &No latent           &Latent           &No latent            \\ \hline
        Positive &0          &0             &0            &0            &0            &0             \\ \hline
        Negative &2         &24             &2          &24            &2           &24             \\ \hline
\end{tabular}
\end{table}

Medium arc strength simulated data:

\begin{table}[H]
\centering
\begin{tabular}{|c|c|c|c|c|c|c|}
\hline
 Data case number& \multicolumn{2}{c|}{100} & \multicolumn{2}{c|}{1000} & \multicolumn{2}{c|}{10000} \\ \hline
                 &Latent         &No latent            &Latent           &No latent           &Latent           &No latent            \\ \hline
        Positive &0          &0             &1            &0            &1            &0             \\ \hline
        Negative &2         &24             &1           &24            &1           &24             \\ \hline
\end{tabular}
\end{table}

Minimum arc strength simulated data:

\begin{table}[H]
\centering
\begin{tabular}{|c|c|c|c|c|c|c|}
\hline
 Data case number& \multicolumn{2}{c|}{100} & \multicolumn{2}{c|}{1000} & \multicolumn{2}{c|}{10000} \\ \hline
                 &Latent         &No latent            &Latent           &No latent           &Latent           &No latent            \\ \hline
        Positive &0          &0             &0            &0            &0            &0             \\ \hline
        Negative &2         &24             &2           &24            &2           &24             \\ \hline
\end{tabular}
\end{table}

\item 4 observed variables with 3 state each simulated data \\

Alpha: 0.20160 \\

Maximum arc strength simulated data:

\begin{table}[H]
\centering
\begin{tabular}{|c|c|c|c|c|c|c|}
\hline
 Data case number& \multicolumn{2}{c|}{100} & \multicolumn{2}{c|}{1000} & \multicolumn{2}{c|}{10000} \\ \hline
                 &Latent         &No latent            &Latent           &No latent           &Latent           &No latent            \\ \hline
        Positive &0          &0             &0            &0            &2            &0             \\ \hline
        Negative &2         &24             &2           &24            &0           &24             \\ \hline
\end{tabular}
\end{table}

Medium arc strength simulated data:

\begin{table}[H]
\centering
\begin{tabular}{|c|c|c|c|c|c|c|}
\hline
 Data case number& \multicolumn{2}{c|}{100} & \multicolumn{2}{c|}{1000} & \multicolumn{2}{c|}{10000} \\ \hline
                 &Latent         &No latent            &Latent           &No latent           &Latent           &No latent            \\ \hline
        Positive &0          &0             &1            &0            &1            &0             \\ \hline
        Negative &2         &24             &1           &24            &1           &24             \\ \hline
\end{tabular}
\end{table}

Minimum arc strength simulated data:

\begin{table}[H]
\centering
\begin{tabular}{|c|c|c|c|c|c|c|}
\hline
 Data case number& \multicolumn{2}{c|}{100} & \multicolumn{2}{c|}{1000} & \multicolumn{2}{c|}{10000} \\ \hline
                 &Latent         &No latent            &Latent           &No latent           &Latent           &No latent            \\ \hline
        Positive &0          &0             &0            &0            &0            &0             \\ \hline
        Negative &2         &24             &2           &24           &2           &24             \\ \hline
\end{tabular}
\end{table}

\item 5 observed variables with 2 state each simulated data \\

Alpha: 0.20676 \\

Maximum arc strength simulated data:

\begin{table}[H]
\centering
\begin{tabular}{|c|c|c|c|c|c|c|}
\hline
 Data case number& \multicolumn{2}{c|}{100} & \multicolumn{2}{c|}{1000} & \multicolumn{2}{c|}{10000} \\ \hline
                 &Latent         &No latent            &Latent           &No latent           &Latent           &No latent            \\ \hline
        Positive &0          &0             &0            &0            &0            &0             \\ \hline
        Negative &57         &268             & 57          &268            &57           &268             \\ \hline
\end{tabular}
\end{table}

Medium arc strength simulated data:

\begin{table}[H]
\centering
\begin{tabular}{|c|c|c|c|c|c|c|}
\hline
 Data case number& \multicolumn{2}{c|}{100} & \multicolumn{2}{c|}{1000} & \multicolumn{2}{c|}{10000} \\ \hline
                 &Latent         &No latent            &Latent           &No latent           &Latent           &No latent            \\ \hline
        Positive &0          &0             &0            &2            &0            &0             \\ \hline
        Negative &57         & 268            &57           &266            &57           &268             \\ \hline
\end{tabular}
\end{table}

Minimum arc strength simulated data:

\begin{table}[H]
\centering
\begin{tabular}{|c|c|c|c|c|c|c|}
\hline
 Data case number& \multicolumn{2}{c|}{100} & \multicolumn{2}{c|}{1000} & \multicolumn{2}{c|}{10000} \\ \hline
                 &Latent         &No latent            &Latent           &No latent           &Latent           &No latent            \\ \hline
        Positive &0          &0             &0            &0            &0            &0             \\ \hline
        Negative &57         &268             &57           &268            &57           &268             \\ \hline
\end{tabular}
\end{table}

\item 5 observed variables with 3 state each simulated data \\

Alpha: 0.13636 \\

Maximum arc strength simulated data:

\begin{table}[H]
\centering
\begin{tabular}{|c|c|c|c|c|c|c|}
\hline
 Data case number& \multicolumn{2}{c|}{100} & \multicolumn{2}{c|}{1000} & \multicolumn{2}{c|}{10000} \\ \hline
                 &Latent         &No latent            &Latent           &No latent           &Latent           &No latent            \\ \hline
        Positive &0          &0             &1            &1            &13            &0            \\ \hline
        Negative &57        &268             &56           &267            &44           &268             \\ \hline
\end{tabular}
\end{table}

Medium arc strength simulated data:

\begin{table}[H]
\centering
\begin{tabular}{|c|c|c|c|c|c|c|}
\hline
 Data case number& \multicolumn{2}{c|}{100} & \multicolumn{2}{c|}{1000} & \multicolumn{2}{c|}{10000} \\ \hline
                 &Latent         &No latent            &Latent           &No latent           &Latent           &No latent            \\ \hline
        Positive &0          &0             &0            &0            &9            &0             \\ \hline
        Negative &57         &268             &57          &268            &48           &268             \\ \hline
\end{tabular}
\end{table}

Minimum arc strength simulated data:

\begin{table}[H]
\centering
\begin{tabular}{|c|c|c|c|c|c|c|}
\hline
 Data case number& \multicolumn{2}{c|}{100} & \multicolumn{2}{c|}{1000} & \multicolumn{2}{c|}{10000} \\ \hline
                 &Latent         &No latent            &Latent           &No latent           &Latent           &No latent            \\ \hline
        Positive &0         &0             &0            &0            &1            &0             \\ \hline
        Negative &57         &268             &57           &268            &56           &268             \\ \hline
\end{tabular}
\end{table}

\end{itemize}

4) Results of FCI algorithm (with alpha of 0.05)

\begin{itemize}

\item 4 observed variables with 2 state each simulated data \\

Maximum arc strength simulated data:

\begin{table}[H]
\centering
\begin{tabular}{|c|c|c|c|c|c|c|}
\hline
 Data case number& \multicolumn{2}{c|}{100} & \multicolumn{2}{c|}{1000} & \multicolumn{2}{c|}{10000} \\ \hline
                 &Latent         &No latent            &Latent           &No latent           &Latent           &No latent            \\ \hline
        Positive &0          &0             &1            &0            &0            &0             \\ \hline
        Negative &2         &24             &1           &24            &2          &24             \\ \hline
\end{tabular}
\end{table}

Medium arc strength simulated data:

\begin{table}[H]
\centering
\begin{tabular}{|c|c|c|c|c|c|c|}
\hline
 Data case number& \multicolumn{2}{c|}{100} & \multicolumn{2}{c|}{1000} & \multicolumn{2}{c|}{10000} \\ \hline
                 &Latent         &No latent            &Latent           &No latent           &Latent           &No latent            \\ \hline
        Positive &0          &1             &2            &1            &2            &0             \\ \hline
        Negative &2         &23             &0           &23            &0           &24             \\ \hline
\end{tabular}
\end{table}

Minimum arc strength simulated data:

\begin{table}[H]
\centering
\begin{tabular}{|c|c|c|c|c|c|c|}
\hline
 Data case number& \multicolumn{2}{c|}{100} & \multicolumn{2}{c|}{1000} & \multicolumn{2}{c|}{10000} \\ \hline
                 &Latent         &No latent            &Latent           &No latent           &Latent           &No latent            \\ \hline
        Positive &0          &0             &0            &0            &0            &0             \\ \hline
        Negative &2         &24            &2           &24            &2           &24             \\ \hline
\end{tabular}
\end{table}

\item 4 observed variables with 3 state each simulated data \\

Maximum arc strength simulated data:

\begin{table}[H]
\centering
\begin{tabular}{|c|c|c|c|c|c|c|}
\hline
 Data case number& \multicolumn{2}{c|}{100} & \multicolumn{2}{c|}{1000} & \multicolumn{2}{c|}{10000} \\ \hline
                 &Latent         &No latent            &Latent           &No latent           &Latent           &No latent            \\ \hline
        Positive &0          &0             &1            &0            &1            &0             \\ \hline
        Negative &2         &24             &1          &24            &1          &24             \\ \hline
\end{tabular}
\end{table}

Medium arc strength simulated data:

\begin{table}[H]
\centering
\begin{tabular}{|c|c|c|c|c|c|c|}
\hline
 Data case number& \multicolumn{2}{c|}{100} & \multicolumn{2}{c|}{1000} & \multicolumn{2}{c|}{10000} \\ \hline
                 &Latent         &No latent            &Latent           &No latent           &Latent           &No latent            \\ \hline
        Positive &0          &0             &1           &1            &2            &0             \\ \hline
        Negative &2         &24             &1           &23            &0          &24             \\ \hline
\end{tabular}
\end{table}

Minimum arc strength simulated data:

\begin{table}[H]
\centering
\begin{tabular}{|c|c|c|c|c|c|c|}
\hline
 Data case number& \multicolumn{2}{c|}{100} & \multicolumn{2}{c|}{1000} & \multicolumn{2}{c|}{10000} \\ \hline
                 &Latent         &No latent            &Latent           &No latent           &Latent           &No latent            \\ \hline
        Positive &0          &0             &0            &0            &0            &6             \\ \hline
        Negative &2         &24             &2           &24            &2           &18             \\ \hline
\end{tabular}
\end{table}

\item 5 observed variables with 2 state each simulated data \\

Maximum arc strength simulated data:

\begin{table}[H]
\centering
\begin{tabular}{|c|c|c|c|c|c|c|}
\hline
 Data case number& \multicolumn{2}{c|}{100} & \multicolumn{2}{c|}{1000} & \multicolumn{2}{c|}{10000} \\ \hline
                 &Latent         &No latent            &Latent           &No latent           &Latent           &No latent            \\ \hline
        Positive &2          &32             &3            &47            &10            &24             \\ \hline
        Negative &55         &236             &54           &221            &47           &244             \\ \hline
\end{tabular}
\end{table}

Medium arc strength simulated data:

\begin{table}[H]
\centering
\begin{tabular}{|c|c|c|c|c|c|c|}
\hline
 Data case number& \multicolumn{2}{c|}{100} & \multicolumn{2}{c|}{1000} & \multicolumn{2}{c|}{10000} \\ \hline
                 &Latent         &No latent            &Latent           &No latent           &Latent           &No latent            \\ \hline
        Positive &1          &29             &8            &63            &33            &49            \\ \hline
        Negative &56         &239             &49           &205            &24           &219             \\ \hline
\end{tabular}
\end{table}

Minimum arc strength simulated data:

\begin{table}[H]
\centering
\begin{tabular}{|c|c|c|c|c|c|c|}
\hline
 Data case number& \multicolumn{2}{c|}{100} & \multicolumn{2}{c|}{1000} & \multicolumn{2}{c|}{10000} \\ \hline
                 &Latent         &No latent            &Latent           &No latent           &Latent           &No latent            \\ \hline
        Positive &0          &0             &0            &2            &0            &7             \\ \hline
        Negative &57         &268             &57           &266            &57           &261             \\ \hline
\end{tabular}
\end{table}

\item 5 observed variables with 3 state each simulated data \\

Maximum arc strength simulated data:

\begin{table}[H]
\centering
\begin{tabular}{|c|c|c|c|c|c|c|}
\hline
 Data case number& \multicolumn{2}{c|}{100} & \multicolumn{2}{c|}{1000} & \multicolumn{2}{c|}{10000} \\ \hline
                 &Latent         &No latent            &Latent           &No latent           &Latent           &No latent            \\ \hline
        Positive &2          &64             &25            &63            &48            &13             \\ \hline
        Negative &55         &204             &32           &205            &9           &255             \\ \hline
\end{tabular}
\end{table}

Medium arc strength simulated data:

\begin{table}[H]
\centering
\begin{tabular}{|c|c|c|c|c|c|c|}
\hline
 Data case number& \multicolumn{2}{c|}{100} & \multicolumn{2}{c|}{1000} & \multicolumn{2}{c|}{10000} \\ \hline
                 &Latent         &No latent            &Latent           &No latent           &Latent           &No latent            \\ \hline
        Positive &1          &65             &17            &116            &48            &15             \\ \hline
        Negative &56         &203             &40           &152            &9           &253             \\ \hline
\end{tabular}
\end{table}

Minimum arc strength simulated data:

\begin{table}[H]
\centering
\begin{tabular}{|c|c|c|c|c|c|c|}
\hline
 Data case number& \multicolumn{2}{c|}{100} & \multicolumn{2}{c|}{1000} & \multicolumn{2}{c|}{10000} \\ \hline
                 &Latent         &No latent            &Latent           &No latent           &Latent           &No latent            \\ \hline
        Positive &0          &2             &0            &29            &3            &138             \\ \hline
        Negative &57         &266             &57           &239            &54           &130             \\ \hline
\end{tabular}
\end{table}

\end{itemize}

5) Results of PC algorithm (with alpha of 0.05)

\begin{itemize}

\item 4 observed variables with 2 state each simulated data \\

Maximum arc strength simulated data:

\begin{table}[H]
\centering
\begin{tabular}{|c|c|c|c|c|c|c|}
\hline
 Data case number& \multicolumn{2}{c|}{100} & \multicolumn{2}{c|}{1000} & \multicolumn{2}{c|}{10000} \\ \hline
                 &Latent         &No latent            &Latent           &No latent           &Latent           &No latent            \\ \hline
        Positive &0          &0             &1            &0            &0            &0             \\ \hline
        Negative &2         &24             &1           &24            &2           &24             \\ \hline
\end{tabular}
\end{table}

Medium arc strength simulated data:

\begin{table}[H]
\centering
\begin{tabular}{|c|c|c|c|c|c|c|}
\hline
 Data case number& \multicolumn{2}{c|}{100} & \multicolumn{2}{c|}{1000} & \multicolumn{2}{c|}{10000} \\ \hline
                 &Latent         &No latent            &Latent           &No latent           &Latent           &No latent            \\ \hline
        Positive &0          &1             &2            &1            &2            &0             \\ \hline
        Negative &2         &23             &0           &23            &0           &24             \\ \hline
\end{tabular}
\end{table}

Minimum arc strength simulated data:

\begin{table}[H]
\centering
\begin{tabular}{|c|c|c|c|c|c|c|}
\hline
 Data case number& \multicolumn{2}{c|}{100} & \multicolumn{2}{c|}{1000} & \multicolumn{2}{c|}{10000} \\ \hline
                 &Latent         &No latent            &Latent           &No latent           &Latent           &No latent            \\ \hline
        Positive &0          &0             &0            &0            &0            &0             \\ \hline
        Negative &2         &24            &2           &24            &2           &24             \\ \hline
\end{tabular}
\end{table}

\item 4 observed variables with 3 state each simulated data \\

Maximum arc strength simulated data:

\begin{table}[H]
\centering
\begin{tabular}{|c|c|c|c|c|c|c|}
\hline
 Data case number& \multicolumn{2}{c|}{100} & \multicolumn{2}{c|}{1000} & \multicolumn{2}{c|}{10000} \\ \hline
                 &Latent         &No latent            &Latent           &No latent           &Latent           &No latent            \\ \hline
        Positive &0          &0             &1            &0            &1            &0             \\ \hline
        Negative &2        &24            &1           &24            &1           &24             \\ \hline
\end{tabular}
\end{table}

Medium arc strength simulated data:

\begin{table}[H]
\centering
\begin{tabular}{|c|c|c|c|c|c|c|}
\hline
 Data case number& \multicolumn{2}{c|}{100} & \multicolumn{2}{c|}{1000} & \multicolumn{2}{c|}{10000} \\ \hline
                 &Latent         &No latent            &Latent           &No latent           &Latent           &No latent            \\ \hline
        Positive &0          &0             &1            &1            &2            &0             \\ \hline
        Negative &2         &24             &1           &23            &0           &24             \\ \hline
\end{tabular}
\end{table}

Minimum arc strength simulated data:

\begin{table}[H]
\centering
\begin{tabular}{|c|c|c|c|c|c|c|}
\hline
 Data case number& \multicolumn{2}{c|}{100} & \multicolumn{2}{c|}{1000} & \multicolumn{2}{c|}{10000} \\ \hline
                 &Latent         &No latent            &Latent           &No latent           &Latent           &No latent            \\ \hline
        Positive &0          &0             &0            &0            &0            &6             \\ \hline
        Negative &2         &24             &2           &24            &2           &18             \\ \hline
\end{tabular}
\end{table}

\item 5 observed variables with 2 state each simulated data \\

Maximum arc strength simulated data:

\begin{table}[H]
\centering
\begin{tabular}{|c|c|c|c|c|c|c|}
\hline
 Data case number& \multicolumn{2}{c|}{100} & \multicolumn{2}{c|}{1000} & \multicolumn{2}{c|}{10000} \\ \hline
                 &Latent         &No latent            &Latent           &No latent           &Latent           &No latent            \\ \hline
        Positive &1          &25             &4            &35            &10            &12             \\ \hline
        Negative &56         & 243            &53           &233            &47           &256             \\ \hline
\end{tabular}
\end{table}

Medium arc strength simulated data:

\begin{table}[H]
\centering
\begin{tabular}{|c|c|c|c|c|c|c|}
\hline
 Data case number& \multicolumn{2}{c|}{100} & \multicolumn{2}{c|}{1000} & \multicolumn{2}{c|}{10000} \\ \hline
                 &Latent         &No latent            &Latent           &No latent           &Latent           &No latent            \\ \hline
        Positive &1          &18             &7            &47            &32            &33             \\ \hline
        Negative &56         &250             &50           &221            &25           &235             \\ \hline
\end{tabular}
\end{table}

Minimum arc strength simulated data:

\begin{table}[H]
\centering
\begin{tabular}{|c|c|c|c|c|c|c|}
\hline
 Data case number& \multicolumn{2}{c|}{100} & \multicolumn{2}{c|}{1000} & \multicolumn{2}{c|}{10000} \\ \hline
                 &Latent         &No latent            &Latent           &No latent           &Latent           &No latent            \\ \hline
        Positive &0          &0             &0            &2            &0            &6             \\ \hline
        Negative &57        &268             &57           &266            &57           &262             \\ \hline
\end{tabular}
\end{table}

\item 5 observed variables with 3 state each simulated data \\

Maximum arc strength simulated data:

\begin{table}[H]
\centering
\begin{tabular}{|c|c|c|c|c|c|c|}
\hline
 Data case number& \multicolumn{2}{c|}{100} & \multicolumn{2}{c|}{1000} & \multicolumn{2}{c|}{10000} \\ \hline
                 &Latent         &No latent            &Latent           &No latent           &Latent           &No latent            \\ \hline
        Positive &2          &56             &24            &48            &47            &7             \\ \hline
        Negative &55         &212             &33           &220            &10           &261             \\ \hline
\end{tabular}
\end{table}

Medium arc strength simulated data:

\begin{table}[H]
\centering
\begin{tabular}{|c|c|c|c|c|c|c|}
\hline
 Data case number& \multicolumn{2}{c|}{100} & \multicolumn{2}{c|}{1000} & \multicolumn{2}{c|}{10000} \\ \hline
                 &Latent         &No latent            &Latent           &No latent           &Latent           &No latent            \\ \hline
        Positive &0          &56             &16            &95            &48            &9             \\ \hline
        Negative &57         &212             &41           &173            &9           &259             \\ \hline
\end{tabular}
\end{table}

Minimum arc strength simulated data:

\begin{table}[H]
\centering
\begin{tabular}{|c|c|c|c|c|c|c|}
\hline
 Data case number& \multicolumn{2}{c|}{100} & \multicolumn{2}{c|}{1000} & \multicolumn{2}{c|}{10000} \\ \hline
                 &Latent         &No latent            &Latent           &No latent           &Latent           &No latent            \\ \hline
        Positive &0          &2             &0            &28            &3            &127             \\ \hline
        Negative &57         &266             &57           &240            &54           &141             \\ \hline
\end{tabular}
\end{table}

\end{itemize}


6) Results of trigger-PC algorithm (with alpha of 0.05)

\begin{itemize}

\item 4 observed variables with 2 state each simulated data \\

Maximum arc strength simulated data:

\begin{table}[H]
\centering
\begin{tabular}{|c|c|c|c|c|c|c|}
\hline
 Data case number& \multicolumn{2}{c|}{100} & \multicolumn{2}{c|}{1000} & \multicolumn{2}{c|}{10000} \\ \hline
                 &Latent         &No latent            &Latent           &No latent           &Latent           &No latent            \\ \hline
        Positive &0          &0             &0            &0            &0            &0             \\ \hline
        Negative &2         &24             &2           &24            &2          &24             \\ \hline
\end{tabular}
\end{table}

Medium arc strength simulated data:

\begin{table}[H]
\centering
\begin{tabular}{|c|c|c|c|c|c|c|}
\hline
 Data case number& \multicolumn{2}{c|}{100} & \multicolumn{2}{c|}{1000} & \multicolumn{2}{c|}{10000} \\ \hline
                 &Latent         &No latent            &Latent           &No latent           &Latent           &No latent            \\ \hline
        Positive &0          &0             &1            &0            &2            &0             \\ \hline
        Negative &2         &24             &1           &24            &0           &24             \\ \hline
\end{tabular}
\end{table}

Minimum arc strength simulated data:

\begin{table}[H]
\centering
\begin{tabular}{|c|c|c|c|c|c|c|}
\hline
 Data case number& \multicolumn{2}{c|}{100} & \multicolumn{2}{c|}{1000} & \multicolumn{2}{c|}{10000} \\ \hline
                 &Latent         &No latent            &Latent           &No latent           &Latent           &No latent            \\ \hline
        Positive &0          &0             &0            &0            &0            &0             \\ \hline
        Negative &2         &24             &2           &24            &2           &24             \\ \hline
\end{tabular}
\end{table}

\item 4 observed variables with 3 state each simulated data \\

Maximum arc strength simulated data:

\begin{table}[H]
\centering
\begin{tabular}{|c|c|c|c|c|c|c|}
\hline
 Data case number& \multicolumn{2}{c|}{100} & \multicolumn{2}{c|}{1000} & \multicolumn{2}{c|}{10000} \\ \hline
                 &Latent         &No latent            &Latent           &No latent           &Latent           &No latent            \\ \hline
        Positive &0          &0             &0            &0            &0            &0             \\ \hline
        Negative &2         &24             &2          &24            &2           &24             \\ \hline
\end{tabular}
\end{table}

Medium arc strength simulated data:

\begin{table}[H]
\centering
\begin{tabular}{|c|c|c|c|c|c|c|}
\hline
 Data case number& \multicolumn{2}{c|}{100} & \multicolumn{2}{c|}{1000} & \multicolumn{2}{c|}{10000} \\ \hline
                 &Latent         &No latent            &Latent           &No latent           &Latent           &No latent            \\ \hline
        Positive &0          &0             &0            &0            &2            &0             \\ \hline
        Negative &2         &24             &2          &24            &0           &24             \\ \hline
\end{tabular}
\end{table}

Minimum arc strength simulated data:

\begin{table}[H]
\centering
\begin{tabular}{|c|c|c|c|c|c|c|}
\hline
 Data case number& \multicolumn{2}{c|}{100} & \multicolumn{2}{c|}{1000} & \multicolumn{2}{c|}{10000} \\ \hline
                 &Latent         &No latent            &Latent           &No latent           &Latent           &No latent            \\ \hline
        Positive &0          &0             &0           &0            &0            &0             \\ \hline
        Negative &2         &24             &2           &24            &2           &24             \\ \hline
\end{tabular}
\end{table}

\item 5 observed variables with 2 state each simulated data \\

Maximum arc strength simulated data:

\begin{table}[H]
\centering
\begin{tabular}{|c|c|c|c|c|c|c|}
\hline
 Data case number& \multicolumn{2}{c|}{100} & \multicolumn{2}{c|}{1000} & \multicolumn{2}{c|}{10000} \\ \hline
                 &Latent         &No latent            &Latent           &No latent           &Latent           &No latent            \\ \hline
        Positive &0          &0             &0            &0            &0            &1             \\ \hline
        Negative &57         &268             &57          &268            &57           &267             \\ \hline
\end{tabular}
\end{table}

Medium arc strength simulated data:

\begin{table}[H]
\centering
\begin{tabular}{|c|c|c|c|c|c|c|}
\hline
 Data case number& \multicolumn{2}{c|}{100} & \multicolumn{2}{c|}{1000} & \multicolumn{2}{c|}{10000} \\ \hline
                 &Latent         &No latent            &Latent           &No latent           &Latent           &No latent            \\ \hline
        Positive &0          &0             &0            &1            &3            &0             \\ \hline
        Negative &57         &268             &57           &267            &54           &268             \\ \hline
\end{tabular}
\end{table}

Minimum arc strength simulated data:

\begin{table}[H]
\centering
\begin{tabular}{|c|c|c|c|c|c|c|}
\hline
 Data case number& \multicolumn{2}{c|}{100} & \multicolumn{2}{c|}{1000} & \multicolumn{2}{c|}{10000} \\ \hline
                 &Latent         &No latent            &Latent           &No latent           &Latent           &No latent            \\ \hline
        Positive &0          &0             &0            &0            &0            &0            \\ \hline
        Negative &57         &268             &57           &268            &57           &268             \\ \hline
\end{tabular}
\end{table}

\item 5 observed variables with 3 state each simulated data \\

Maximum arc strength simulated data:

\begin{table}[H]
\centering
\begin{tabular}{|c|c|c|c|c|c|c|}
\hline
 Data case number& \multicolumn{2}{c|}{100} & \multicolumn{2}{c|}{1000} & \multicolumn{2}{c|}{10000} \\ \hline
                 &Latent         &No latent            &Latent           &No latent           &Latent           &No latent            \\ \hline
        Positive &0          &0             &0            &1            &14            &0             \\ \hline
        Negative &57         &268             &57           &267            &43           &268             \\ \hline
\end{tabular}
\end{table}

Medium arc strength simulated data:

\begin{table}[H]
\centering
\begin{tabular}{|c|c|c|c|c|c|c|}
\hline
 Data case number& \multicolumn{2}{c|}{100} & \multicolumn{2}{c|}{1000} & \multicolumn{2}{c|}{10000} \\ \hline
                 &Latent         &No latent            &Latent           &No latent           &Latent           &No latent            \\ \hline
        Positive &0          &0             &0            &1            &12            &0             \\ \hline
        Negative &57         &268             &57           &267            &45           &268             \\ \hline
\end{tabular}
\end{table}

Minimum arc strength simulated data:

\begin{table}[H]
\centering
\begin{tabular}{|c|c|c|c|c|c|c|}
\hline
 Data case number& \multicolumn{2}{c|}{100} & \multicolumn{2}{c|}{1000} & \multicolumn{2}{c|}{10000} \\ \hline
                 &Latent         &No latent            &Latent           &No latent           &Latent           &No latent            \\ \hline
        Positive &0          &0             &0            &0            &1            &0             \\ \hline
        Negative &57         &268             &57           &268            &56           &268             \\ \hline
\end{tabular}
\end{table}

\end{itemize}


\begin{thebibliography}{1}

\bibitem{friedman1997learning} Friedman, N. (1997). Learning belief
  networks in the presence of missing values and hidden variables
  (pp. 125-133). Int. Conf. on Machine
  Learning.

\bibitem{neticaAPI} Netica API. {\tt https://www.norsys.com/netica\_api.html}

\bibitem{neticasampling} Netica API. {\tt https://www.norsys.com/netica-j/docs/\\javadocs/norsys/netica/Net.html\#FORWARD\_SAMPLING}

\bibitem{pearl1988probabilistic} Pearl, J. (1988). Probabilistic
  Reasoning in Intelligent Systems: Networks of Plausible
  Inference. Morgan Kaufmann.
  
\bibitem{Russel2003} Russel, S. \& Norvig, P. (2003). Artificial Intelligence: A Modern Approach. EUA: Prentice Hall.

\bibitem{robinson1977counting} Robinson, R. W. (1977). Counting
  unlabeled acyclic digraphs. Comb. Math. V
  (pp. 28-43). Springer Verlag.

\bibitem{Spear1904}
Spearman, C. (1904). General intelligence, objectively determined and
measured. The Amer. Jrn. of Psychology, 15(2), 201--292.

\bibitem{spirtes2000causation} Spirtes, P., Glymour, C. N., \&
  Scheines, R. (2000). Causation, Prediction, and Search. MIT press.

\bibitem{tetrad} TETRAD V. {\tt http://www.phil.cmu.edu/projects/tetrad/\\current.html}

\end{thebibliography}
\end{document}